\documentclass[runningheads]{llncs}
\usepackage[T1]{fontenc}
\usepackage{amsmath}
\usepackage{amssymb}
\usepackage{bbm}
\usepackage{graphicx,verbatim}
\usepackage{booktabs} 
\usepackage{multirow}
\usepackage[table]{xcolor}
\usepackage{color}
\usepackage{subfigure}
\usepackage{arydshln}
\usepackage{marvosym}
\usepackage{hyperref}
\usepackage{textcomp}
\usepackage{color}

\begin{document}
\title{Uncertainty-Supervised Interpretable and Robust Evidential Segmentation}
\titlerunning{USIRES}
%

\makeatletter
\renewcommand\@fnsymbol[1]{%
  \ensuremath{%
    \ifcase#1\or
      *\or 
      \dagger\or 
      \ddagger\or 
      \S\or 
      \text{\Letter}\or 
      \mathparagraph\or 
      \|\or 
      **\or 
      \dagger\dagger 
    \fi
  }%
}
\makeatother

\renewcommand{\thefootnote}{\fnsymbol{footnote}}

\author{Yuzhu Li\inst{1, 2}\protect\footnotemark[2] \and
An Sui\inst{1}\protect\footnotemark[2] \and
Fuping Wu\inst{3}\protect\footnotemark[5] \and
Xiahai Zhuang\inst{1}\protect\footnotemark[5]}
\authorrunning{Y. Li et al.}
%
\institute{School of Data Science, Fudan University, Shanghai, China 
\and Institute of Science and Technology for Brain-inspired intelligence, Fudan University, Shanghai, China
\and
Nuffield Department of Population Health, University of Oxford, Oxford, UK
\\\email{zxh@fudan.edu.cn}
\\\url{https://zmiclab.github.io/}}
\maketitle              
\footnotetext[2]{These two authors contributed equally.}
\footnotetext[5]{Xiahai Zhuang and Fuping Wu are the corresponding authors.}
\begin{abstract}
Uncertainty estimation has been widely studied in medical image segmentation as a tool to provide reliability, particularly in deep learning approaches.
However, previous methods generally lack effective supervision in uncertainty estimation, leading to low interpretability and robustness of the predictions. 
In this work, we propose a self-supervised approach to guide the learning of uncertainty.
Specifically, we introduce three principles about the relationships between the uncertainty and the image gradients around boundaries and noise.
Based on these principles, two uncertainty supervision losses are designed.
These losses enhance the alignment between model predictions and human interpretation. 
Accordingly, we introduce novel quantitative metrics for evaluating the interpretability and robustness of uncertainty. Experimental results demonstrate that compared to state-of-the-art approaches, the proposed method can achieve competitive segmentation performance and superior results in out-of-distribution (OOD) scenarios while significantly improving the interpretability and robustness of uncertainty estimation.
Code is available via \url{https://github.com/suiannaius/SURE}.

\end{abstract}
%
%
%
\section{Introduction}\label{sec:introduction}

Accurate medical image segmentation is essential for clinical applications such as diagnosis \cite{addimulam2020deep} and treatment planning \cite{erdur2024deep}. Beyond accuracy, reliability, interpretability, and robustness have raised increasing concerns for researchers and clinicians. 
Recent advances in deep learning, particularly U-Net \cite{falk2019u} and its variants \cite{krithika2022review,huang2020unet,cao2022swin}, have significantly improved segmentation accuracy. However, these models often neglect uncertainty in ambiguous regions like low-contrast or noisy areas, leading to over-confident predictions and errors. The absence of uncertainty further limits access to reliable predictions, hindering practical utility.

To address these issues, uncertainty estimation methods, such as Bayesian approaches \cite{gal2016dropout}, ensemble strategies \cite{lakshminarayanan2017simple}, test time augmentation (TTA) \cite{wang2019aleatoric}, and evidential deep learning (EDL) \cite{huang2022lymphoma}, have emerged. While Bayesian methods like Monte Carlo Dropout \cite{yu2019uncertainty} are computationally costly, ensemble-based techniques \cite{mehrtash2020confidence} require training multiple models, and TTA depends heavily on augmentations, EDL \cite{sensoy2018evidential} offers a computationally efficient and theoretically sound solution. Based on Dempster-Shafer theory \cite{shafer1992dempster} and subjective logic \cite{josang2016subjective}, EDL integrates uncertainty directly into the model, enabling reliable estimations with a single forward pass.
While these models can provide pixel-level confidence \cite{ovadia2019can}, they often fail to explain the underlying mechanism for uncertainty or maintain robustness under noise perturbations. 

In this study, we propose a self-supervised approach to enhance the uncertainty interpretability and robustness against noise based on EDL. 
Different from uncertainty calibration, differentiating the inaccurate predictions from the accurate \cite{zou2023towards}, 
we propose three principles requiring uncertainty estimation conforming to human beings' thinking or reasoning patterns.
Based on these principles, we design supervision losses accordingly, leading to our novel uncertainty supervision learning framework for medical image segmentation.

The contributions of this work are summarized as follows: (1) We introduce an uncertainty supervision approach to enhance the interpretability and robustness of evidential learning, by regularizing the relationships of uncertainty with gradients of boundaries and noise; (2) We introduced new quantitative metrics for interpretability and noise robustness of uncertainty; (3) Experimental results show that the proposed method aligns with human logic and demonstrate enhanced robustness against noise in Out-Of-Distribution (OOD) cases.

\section{Methods}\label{sec:methods}
\begin{figure}[ht]
    \centering
    \includegraphics[width=0.7\linewidth]{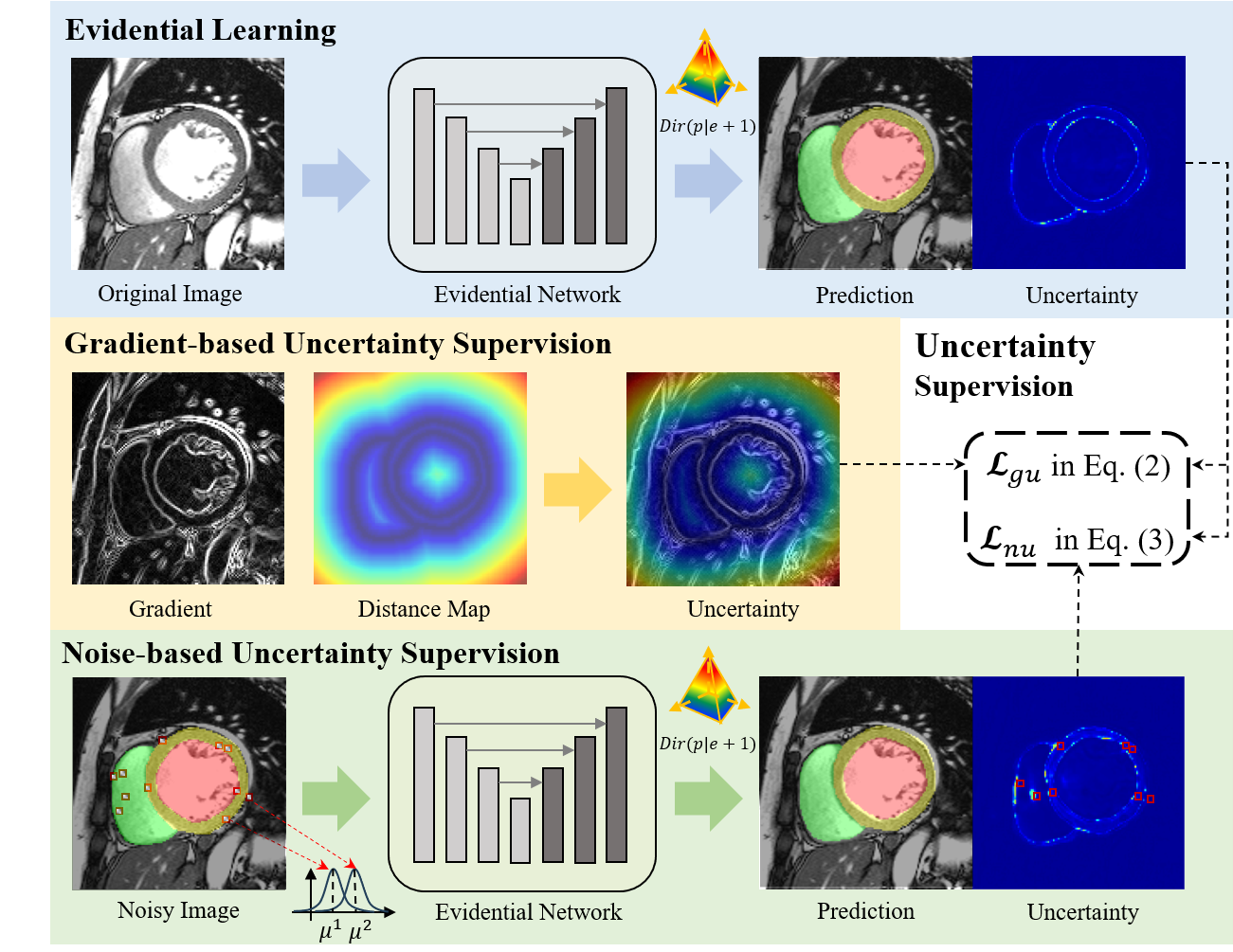}
    \caption{The overview of our work. Based on EDL, our model estimates uncertainty from evidence,  with gradient-based and noise-based supervision to enhance both interpretability and robustness of uncertainty.}
    \label{fig:overview}
\end{figure}

As illustrated in Fig. \ref{fig:overview}, our framework comprises three parts:(1) For segmentation prediction and uncertainty estimation, we employ EDL to generate class-specific evidence for input images, as detailed in Section \ref{subsec:evidential}. (2) For uncertainty supervision, we introduce human-inspired gradient-based supervision loss to enhance its interpretability in Section \ref{subsec:gs}, and (3) we design novel noise-based supervision loss to improve both interpretability and robustness in Section \ref{subsec:ns}. 

\subsection{Evidence-Based Prediction Generation}\label{subsec:evidential}
Let $\boldsymbol{X}=\left(\boldsymbol{x}_i\right)\in\mathbb{R}^V$ and $\boldsymbol{Y}=\left (\boldsymbol{y}_i\right ) \in \mathbb{R}^ {V\times K}$ respectively denote a 2D slice and its label, where $V$ is the number of pixels, $K$ is the number of classes.
The evidential network $f_\theta$ estimates the evidence map as $\boldsymbol{E}=f_\theta(\boldsymbol{X})=\left(\boldsymbol{e}_i\right)\in\mathbb{R}^{V\times K}$.
According to subjective logic \cite{josang2016subjective}, for the $i$-th pixel in $\boldsymbol{X}$, its categorical probability variable $\boldsymbol{p}_i=\left(p_{ik} \right)\in[0,1]^K$ can be modeled as the Dirichlet distribution. 
The uncertainty of the $i$-th pixel can be derived as $u_i=\frac{K}{\sum_{j=1}^K(\boldsymbol{e}_{ij}+1)}$.

For supervised learning, we adopt Dice and cross-entropy loss function, denoted by $\mathcal{L}_{Dice}\left(\boldsymbol{P},\boldsymbol{Y}\right)$ and $\mathcal{L}_{CE}\left(\boldsymbol{P},\boldsymbol{Y}\right)$, proposed in \cite{zou2023evidencecap}, and the Kullback-Leibler divergence, \textit{i.e.,} $\mathcal{L}_{KL}(\boldsymbol{P})$, proposed in \cite{sensoy2018evidential} to avoid collecting evidence about the incorrect classes, where $\boldsymbol{P}=(\boldsymbol{p}_i)\in\mathbb{R}^{V\times K}$. 
The overall EDL segmentation loss is denoted as:
\begin{equation}
    \mathcal{L}_{Seg}=\lambda_{CE}\cdot\mathcal{L}_{CE}+\lambda_{Dice}\cdot\mathcal{L}_{Dice} + \lambda_{KL}\cdot\mathcal{L}_{KL}.
    \label{eq:seg_loss}
\end{equation}

Although evidential learning offers a computationally efficient framework for uncertainty estimation, it lacks supervision regarding the underlying originality, resulting in limited interpretability and robustness. To address this, we propose two uncertainty supervision techniques: gradient-based supervision and noise-based supervision.

\subsection{Gradient-Based Supervision}\label{subsec:gs}

In regions near the boundaries, where uncertainty tends to be high, previous models fail to explicitly reveal the factors contributing to the uncertainty values. As a result, understanding the nature and origin of uncertainty in decision-making processes becomes challenging.
To this end, we propose the following principle to align uncertainty estimation with human intuition:
\\
\\
\textbf{Principle 1.} \textit{For clear boundaries, higher gradients yield lower uncertainty, whereas ambiguous boundaries with lower gradients have higher uncertainty. }
\\

To achieve this principle, we introduce a gradient-based uncertainty supervision loss on boundary pixels, defined as:

\begin{equation}
    \mathcal{L}_{gu} = \frac{1}{|B|} \sum_{i,j \in B, i \neq j} \max(0, (u_i-u_j) (g_i-g_j)),
    \label{eq:gu_loss}
\end{equation}
where \(B\) represents the set of boundary pixels, and \(g_i\) denotes the gradient of pixel $i$, which is computed on a Gaussian smoothed image for stability. 
This formulation ensures that uncertainty estimations are aligned with human perceptual logic, thus enhancing their interpretability.

\subsection{Noise-Based Supervision}\label{subsec:ns}

Uncertainty is intrinsically related to both the noise and the distance of the pixel from the boundary. The following principles describe the relationships:
\\
\\
\textbf{Principle 2.} \textit{When a pixel is close to the boundary, a larger noise amplitude leads to higher uncertainty, and vice versa.}
\\
\\
\textbf{Principle 3.} \textit{When a pixel is sufficiently far from the boundary, the uncertainty becomes negligible, regardless of the noise amplitudes.}
\\
\\

To capture the relationships described in \textbf{Principle 2} and \textbf{Principle 3}, we formulate the noise supervision loss function $\mathcal{L}_{nu}$ as follows for more interpretable and robust uncertainty estimation,

\begin{equation}
    \mathcal{L}_{nu}=\sum_{i\in S} \underbrace{\mathbbm{1}_{d_i \leq d_0} \cdot \max(0, -(\mu^2 - \mu^1)(u_{i}^2 - u_{i}^1))}_{\mathit{nearby} \ \mathit{noise}  \ \rightarrow{}\ \mathit{interpretability}}+\underbrace{\mathbbm{1}_{d_i>d_0} \cdot (u_{i}^0 + u_{i}^1 + u_{i}^2)}_{\mathit{remote} \ \mathit{noise} \rightarrow{}\ \mathit{robustness}},
    \label{eq:nu_loss}
\end{equation}

where $S$ denotes the sampled pixel set, $d_i$ represents the distance of the $i$-th pixel to the boundary, $d_0$ is a threshold, $\mu^1$ and $\mu^2$ denote mean values of two different normal distributions for noise sampling, $u_{i}^1$ and $u_{i}^2$ denote the corresponding uncertainty of the $i$-th pixel after respectively applying the two noises, and $u_{i}^0$ represents its uncertainty without noise.
The first term in the right side of Eq.\eqref{eq:nu_loss} is for interpretability enhancement, and the second term improves the robustness by constraining the uncertainty of pixels with distance larger than $d_0$.

Due to the large number of boundary points, we utilize an active learning strategy that selectively focuses on the most informative data points, \textit{i.e.,} hard samples, to improve training efficiency, instead of using all pixels in Eq. \eqref{eq:nu_loss}. 

\subsubsection{Hard Samples Detection}\label{subsec:hard_samples}
Inspired by multi-class active learning techniques \cite{joshi2009multi}, we first identify hard samples to guide the model's learning process, ensuring it focuses on more challenging instances. Specifically, we impose noise to the entire training images and feed them into the model to obtain noised uncertainty $u^1$. 
According to \textbf{Principle 2}, $u^1$ should be larger than the original uncertainty $u^0$ for all pixels. 
We define hard samples as those pixels not meeting this condition, namely $S^{hard}=\{i|u^1_i\leq u^0_i\}$.

To mitigate class imbalance, 
we adopt a class-wise sampling strategy, which equally samples pixels from the previously identified hard samples for each class. Thus, the model is encouraged to allocate balanced attention to each class region during training, thereby improving its ability to handle underrepresented classes.

\subsection{Total Loss}

In general, the total loss consists of three terms, \textit{i.e.}, the segmentation loss from EDL given by Eq. \eqref{eq:seg_loss}, and the uncertainty supervision losses outlined in Eqs. \eqref{eq:gu_loss} and \eqref{eq:nu_loss},

\begin{equation}
    \mathcal{L}_{total} = \mathcal{L}_{Seg} + {\beta\cdot\mathcal{L}_{gu} + \gamma\cdot\mathcal{L}_{nu}}.
\end{equation}

\section{Experiments}\label{sec:experiments}

\subsection{Dataset and Experiment Setting}\label{subsec:exp-settings}
We validated the proposed method with two datasets:
(1) The \textbf{Automated Cardiac Diagnosis Challenge (ACDC)} dataset contains 200 annotated short-axis cardiac MR-cine images from 100 patients \cite{bernard2018deep}. All slices were cropped to a size of 96 $\times$ 96. 
(2) The \textbf{REFUGE} dataset includes 400 color fundus photography (CFP) images for training and an additional 400 images for testing \cite{ORLANDO2020101570}. Each image was annotated with optic cup (OC) and optic disc (OD) labels. All images were cropped into 512 $\times$ 512.

\textbf{Implementation Details:} We employed U-Net as the backbone, and the Adam optimizer with a learning rate of 0.001. The batch size was set to 24 for ACDC and 8 for REFUGE (reduced to 1 for the PU \cite{eslami2018probabilistic} method due to GPU memory constraint). 
For hyper-parameters, we set $\lambda_{CE}=1, \lambda_{KL}=min(1,t/20)$. We set $\lambda_{Dice}=1-\alpha$ and $\beta=0.1\alpha, \gamma=10\alpha$ for ACDC, and $\beta=\gamma=\alpha$ for REFUGE, where
 the annealing factor $\alpha=\alpha_{0}e^{\{-(\mathrm{In}\alpha_{0}/T)t\}}$. $T$ and $t$ were the total epochs and the current epoch, respectively, with $\alpha_0= 0.01$.
The boundary set $B$ included pixels with $d \leq 1$. For noise supervision, we set $d_0=4$. All experiments were implemented on an NVIDIA Geforce RTX 2080Ti GPU.

\subsection{Evaluation Metrics for \textbf{Principle 1} and \textbf{Principle 2}}\label{subsec:exp-metrics}

For quantitative evaluation of uncertainty, the conventional metrics such as Expected Calibration Error (ECE) and Uncertainty-Error Overlap (UEO) \cite{guo2017calibration,jungo2019assessing} can not measure the interpretable factors proposed in \textbf{Principle 1} and \textbf{Principle 2}.
Therefore, we introduce two sets of new metrics, including Uncertainty Correlation Coefficient (UCC) and Uncertainty Ratio (UR), to quantify the interpretability of uncertainty estimations.

For \textbf{Principle 1}, UCC is defined as the Spearman correlation coefficients \cite{wissler1905spearman} between image gradients and the uncertainty estimations, denoted by $UCC_{[g]}=SCorr(g,u)$ for boundary pixels $B$.
Similarly, for \textbf{Principle 2}, we define $UCC_{[\mu]}=SCorr(\mu,u)$ using pixels with $d \leq d_0$. Take $UCC_{[g]}$ as an example,

\begin{equation}
    UCC_{[g]} = \frac{\sum_{i \in B} (R(g_i)-\overline{R(g)})(R(u_i)-\overline{R(u)})}{\sqrt{\sum_{i \in B} (R(g_i)-\overline{R(g)})^2 \sum_{i \in B} (R(u_i)-\overline{R(u)})^2}} \\.
\end{equation}
$R(\cdot)$ denotes the ranking function, $\overline{(\cdot)}$ denotes the mean values.

The UCC value ranges from [-1, 1], where the sign indicates the direction of correlation (positive or negative), and the magnitude reflects its strength. An interpretable uncertainty should satisfy $UCC_{[g]}<0$ and $UCC_{[\mu]}>0$.

Alternatively, UR calculates the ratio of pixel pairs satisfying the relationships between image gradients (noise) and the uncertainty estimations.
Thus for \textbf{Principle 1}, we define
\begin{equation}
    UR_{[g]}=\frac{\sum_{i,j \in B, i\neq j}\mathbbm{1}_{((g_i-g_j)(u_i-u_j)\leq 0)}}{\sum_{i,j \in B}\mathbbm{1}_{(i\neq j)}}.
\end{equation}
Similarly, we have $UR_{[\mu]}$ for \textbf{Principle 2} defined in pixels with $d\leq d_0$.

For segmentation accuracy, we adopted the Dice Similarity Coefficient
(DSC) and the 95\% Hausdorff Distance (HD95)\cite{huttenlocher1993comparing} as metrics. 

\begin{table}[t]
    \centering
    \caption{Comparisons with various uncertainty estimation methods on the ACDC and REFUGE dataset. The \textbf{bold} indicates the best result in a column, and the \underline{underlined} indicates the second best. (\checkmark) and ($\times$) in UCC columns denote the same and opposite signs as expected, respectively. The unit of HD95 is mm for ACDC, pixels for REFUGE.}
    \begin{tabular}{ccccccccc} 
         \toprule
         \multirow{2}{*}{Methods} & {DSC$\uparrow$} & {HD95$\downarrow$} & \multirow{2}{*}{UEO$\uparrow$ } & \multirow{2}{*}{ECE$\downarrow$ } & \multicolumn{2}{c}{UCC} & \multicolumn{2}{c}{UR$\uparrow$ } \\ 
         \cmidrule(lr{0pt}){6-7} \cmidrule(lr{0pt}){8-9}
         & {(\%)} & {(mm/pixels)} & & & {$g(-)$} & {$\mu(+)$} & {$g$} & {$\mu$} \\ 
         \midrule
         \multicolumn{9}{c}{\textbf{ACDC}} \\
         \midrule
         {\textbf{Ours}} & \textbf{91.06} & {8.45} & {0.222} & {0.009} & {-0.427(\checkmark)} & {0.170(\checkmark)} & \textbf{0.632} & {0.585} \\
         {\textbf{DEviS}} & \underline{90.36} & \underline{7.42} & \underline{0.269} & \textbf{0.007} & {0.109($\times$)} &{-0.002($\times$)} & {0.446} & {0.500} \\
         {\textbf{PU}} & {87.35} & {9.45} & {0.177} & {0.011} & {0.162($\times$)} & {0.602(\checkmark)} & {0.430} & \underline{0.801} \\
         {\textbf{EU}} & {88.14} & \textbf{7.11} & {0.246} & \underline{0.008} & {0.180($\times$)} & {0.021(\checkmark)} & {0.424} & {0.511} \\
         {\textbf{UDrop}} & {88.16} & {7.77} & {0.149} & {0.276} & {-0.022(\checkmark)} & {0.636(\checkmark)} & \underline{0.630} & \textbf{0.818} \\
         {\textbf{TTA}} & {73.44} & {37.7} & \textbf{0.277} & {0.025} & {-0.036(\checkmark)} & {-0.093($\times$)} & {0.497} & {0.453} \\
         \midrule
         \multicolumn{9}{c}{\textbf{REFUGE}} \\
         \midrule
         \textbf{Ours} & \textbf{84.46} & \underline{56.35} & {0.275} & \textbf{0.024} & {-0.056(\checkmark)} & {0.064(\checkmark)} & \textbf{0.519} & {0.532}\\
         \textbf{DEviS} & {83.05} & {65.39} & \underline{0.359} & {0.065} & {0.043($\times$)} & {0.150(\checkmark)} & {0.486} & \underline{0.575}\\
         \textbf{PU} & {79.01} & {117.7} & \textbf{0.384} & \underline{0.035} & {-0.044(\checkmark)} & {0.106(\checkmark)} & {0.515} & {0.553}\\
         \textbf{EU} & \underline{83.60} & \textbf{56.14} & {0.160} & {0.037} & {-0.034(\checkmark)} & {0.078(\checkmark)}  & {0.512} & {0.539} \\
         \textbf{UDrop} & {73.60} & {65.23} & {0.117} & {0.277} & {0.016($\times$)} & {0.110(\checkmark)} & {0.497} & {0.555}\\
         \textbf{TTA} & {75.34} & {98.99} & {0.305} & {0.051} & {-0.052(\checkmark)} & {0.155(\checkmark)} & \underline{0.517} & \textbf{0.578} \\

         \bottomrule
         
    \end{tabular}
    \label{tab:model_comparison_refuge}
\end{table}

\begin{table}[t]
    \centering
    \caption{Ablation study on the REFUGE dataset.}
    \begin{tabular}{ccccccccccc} 
         \toprule
         \multirow{2}{*}{\textbf{$\mathcal{L}_{gu}$}} & \multirow{2}{*}{\textbf{$\mathcal{L}_{nu}$}} & \multirow{2}{*}{$\mathrm{HSD}$} & {DSC$\uparrow$} & {HD95$\downarrow$} & \multirow{2}{*}{UEO$\uparrow$} & \multirow{2}{*}{ECE$\downarrow$} & \multicolumn{2}{c}{UCC} & \multicolumn{2}{c}{UR$\uparrow$} \\ 
         \cmidrule(lr{0pt}){8-9} \cmidrule(lr{0pt}){10-11}
         & & & {(\%)} & {(pixels)} & & & {$g(-)$} & {$\mu(+)$} & {$g$} & {$\mu$}  \\ 
         \midrule
         
         \checkmark & \checkmark & \checkmark & {84.46} & {56.35} & {0.275} & {0.024} & {-0.056(\checkmark)} & {0.064(\checkmark)} & {0.519} & \textbf{0.532}\\
         \cdashline{1-11}
         {$\times$} & \checkmark & \checkmark & \underline{84.51} & \underline{41.27} & \textbf{0.338} & \underline{0.017} & {-0.034(\checkmark)} & {-0.013($\times$)} & {0.511} & {0.493} \\
         \checkmark & {$\times$} & \checkmark & \textbf{84.65} & {42.53} & {0.286} & {0.019} &{-0.064(\checkmark)} & {-0.073($\times$)} & \underline{0.522} & {0.464} \\
         \checkmark & \checkmark & {$\times$} & {83.24} & \textbf{40.75} & {0.319} & \textbf{0.016} & {-0.075(\checkmark)} & {0.009(\checkmark)} & \textbf{0.525} & \underline{0.504} \\
         {$\times$} & {$\times$} & {$\times$} & {83.96} & {58.40} & \underline{0.337} & {0.022} & {-0.029(\checkmark)} & {-0.033($\times$)} & {0.510} & {0.484} \\

         \bottomrule
         
    \end{tabular}
    \label{tab:model_ablation_refuge}
\end{table}

\begin{figure}[t]
    \centering
    \subfigure[DSC change.]{\includegraphics[width=.49\textwidth]{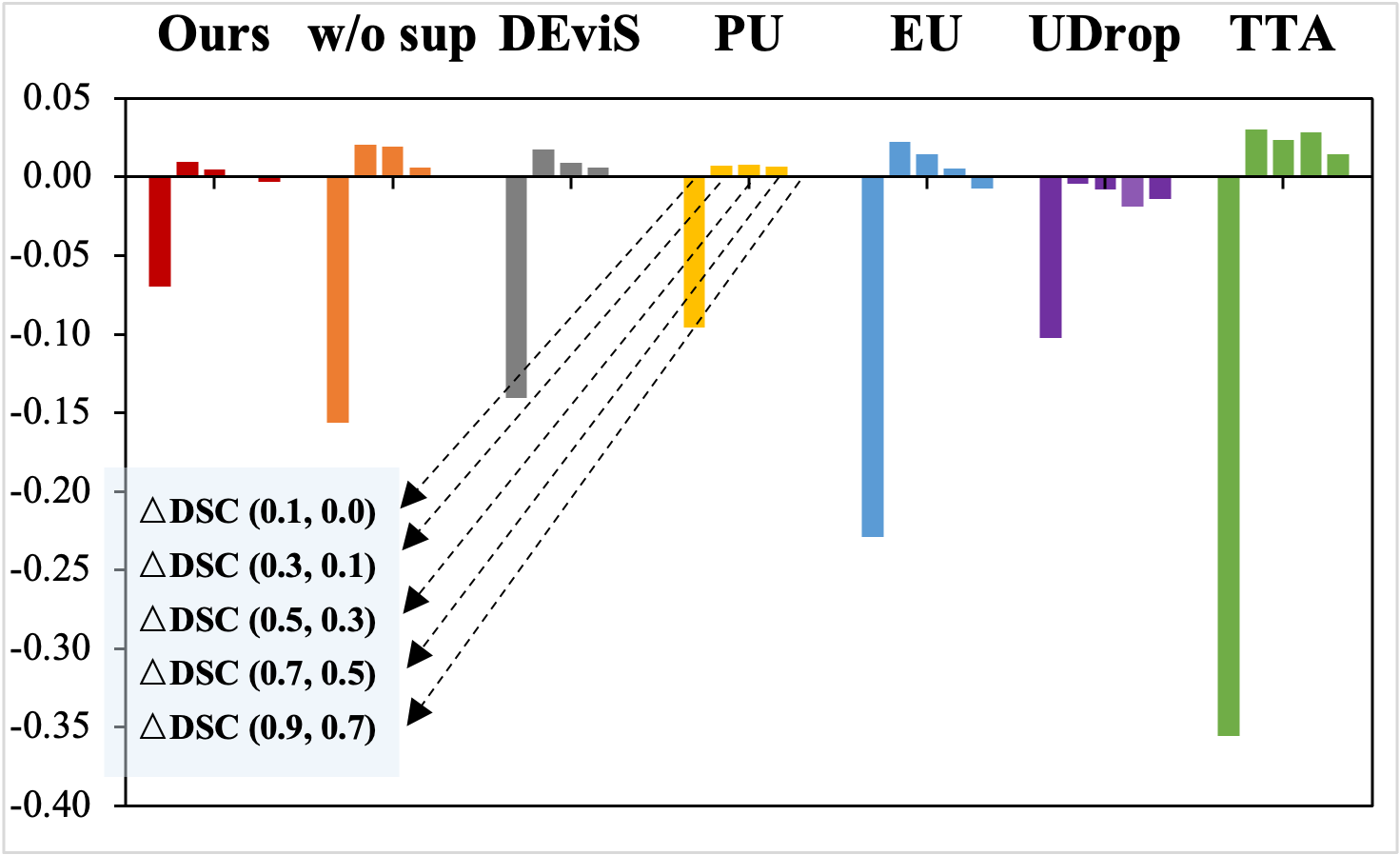}}
    \subfigure[ECE change.]
    {\includegraphics[width=.49\textwidth]{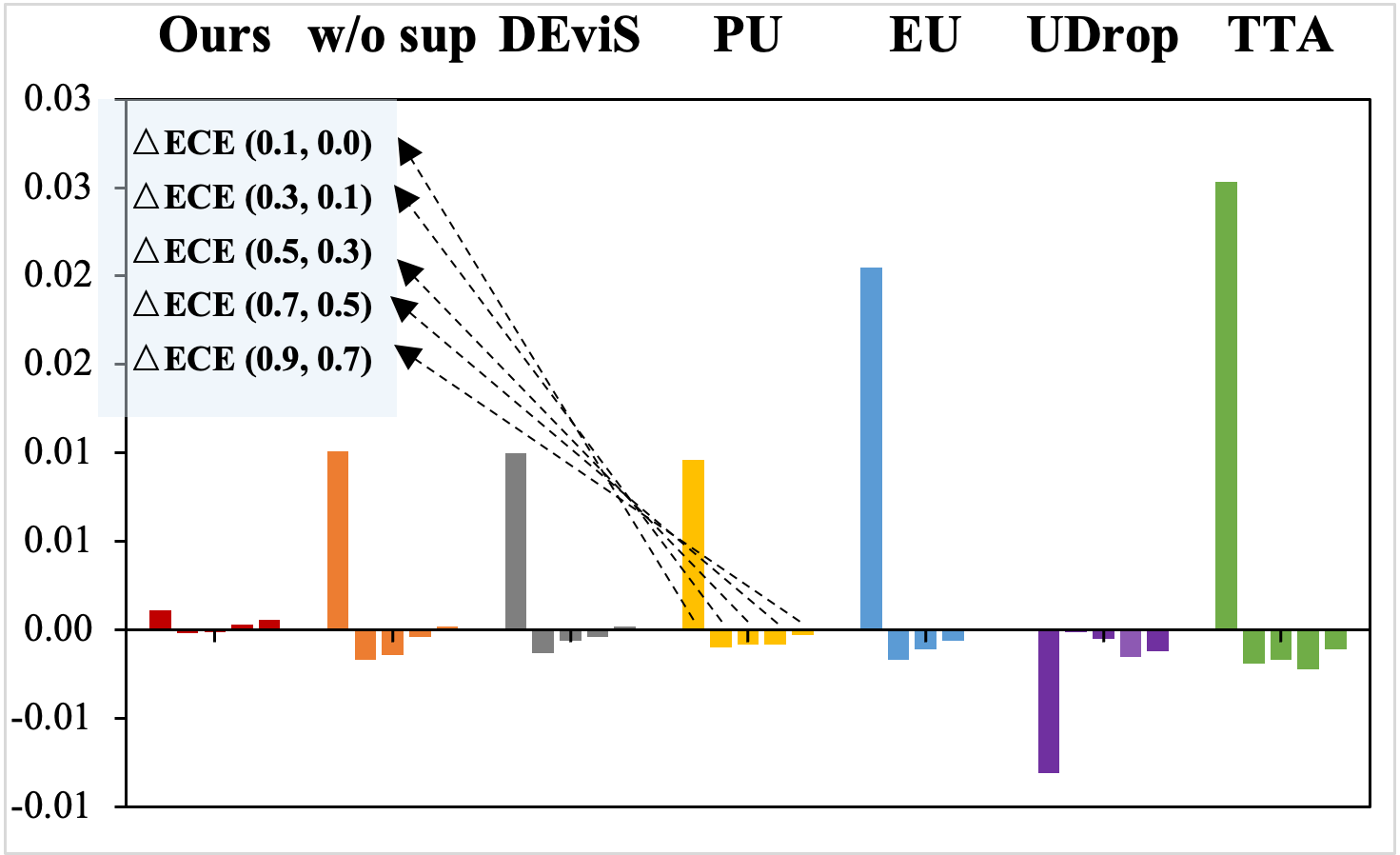}}
    \caption{Robustness evaluation on ACDC dataset, different colors represent different methods. The values are the difference of scores under two different noise levels, \textit{e.g.,} $\triangle DSC(0.3,0.1) = DSC(\mu=0.3)-DSC(\mu=0.1)$.}
    \label{fig:robustness}
\end{figure}
\subsection{Experiment Results}\label{subsec:exp-results}
\subsubsection{Comparison Study} 
We evaluated the proposed method by comparing with various uncertainty estimation approaches, including EDL-based DEviS \cite{zou2023towards}, Variational Inference-based Probabilistic U-Net (PU) \cite{eslami2018probabilistic}, Deep Ensemble-based EU \cite{lakshminarayanan2017simple}, Dropout-based UDrop \cite{kendall2017uncertainties}, and Test-Time Augmentation (TTA) \cite{wang2019aleatoric}.

Table \ref{tab:model_comparison_refuge} presents the quantitative results on ACDC and REFUGE: (1) For \textbf{segmentation accuracy}, our method achieved the best DSC scores and competitive HD95 values on both datasets. (2) For the conventional \textbf{uncertainty evaluation (UEO, ECE)}, our method obtained competitive results on both datasets.
Particularly on REFUGE dataset, we achieved the best ECE value.
Note that UEO measures a strong correlation between uncertainty and error, which might not match our principles to some extent. To enhance the interpretability of uncertainty with the new uncertainty supervision losses, the performance in UEO can be traded off. (3) For \textbf{uncertainty interpretability (UCC, UR)}, our method demonstrated significant superiority, as evidenced by the consistent signs of UCC and their values in the tables.
Moreover, one can see that for other compared methods, none of them obtained the right signs for $UCC_{[g]}$ and $UCC_{[\mu]}$ on both datasets completely.
Specifically, DEviS only had the right sign of $UCC_{[\mu]}$ on REFUGE, both PU and EU obtained the wrong sign of  $UCC_{[g]}$ on ACDC, UDrop and TTA respectively got the wrong sign of $UCC_{[g]}$ on REFUGE and $UCC_{[\mu]}$ on ACDC.


\subsubsection{Ablation Study}
We analyzed the effectiveness of three techniques adopted in the proposed method, including (1) gradient supervision loss $\mathcal{L}_{gu}$ (Eqs.\eqref{eq:gu_loss}), (2) noise supervision loss $\mathcal{L}_{nu}$ (Eqs.\eqref{eq:nu_loss}), 
and (3) hard sample detection (HSD) (Sec.\ref{subsec:hard_samples}). As showed in Table \ref{tab:model_ablation_refuge} , without either $\mathcal{L}_{gu}$ or $\mathcal{L}_{nu}$, our model delivered an opposite sign of $UCC_{[\mu]}$, although other metrics were slightly improved. Note that $\mathrm{HSD}$ was used in the sampling process of the noise supervision loss, its removal also led to a decrease of $UCC_{[\mu]}$ and $UR_{[\mu]}$.

\begin{figure}[t]
    \centering
    \includegraphics[width=1.0\linewidth]{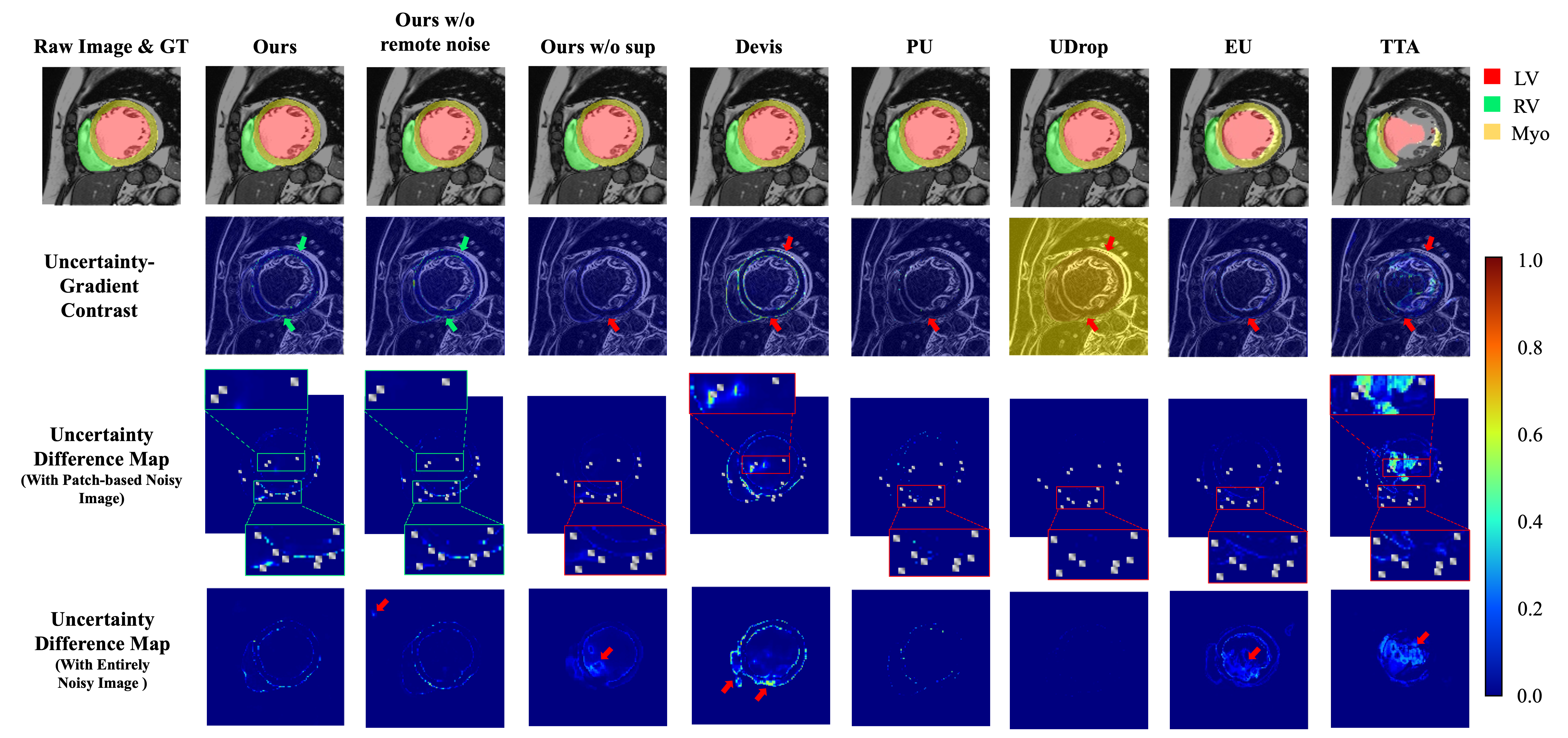}
    \caption{Illustration of prediction results and uncertainty maps of different methods. Red arrows/boxes highlight erroneous uncertainty estimations, while green indicates correct ones.}
    \label{fig:uncertainty_results}
\end{figure}
\subsubsection{Discussion}
To validate \textbf{Principle 3} for robustness enhancing, we evaluated the performance change when exposed to different levels of noise.
Fig. \ref{fig:robustness} (a) illustrates the change of DSC values for all compared methods.
Specifically, we use $\triangle DSC(\mu^i,\mu^j)$ representing the difference of DSC scores when applying two noises with mean value being $\mu_i$ and $\mu_j$ respectively.
We chose $\mu_i \in \{0.0,0.1,0.3,0.5,0.7,0.9\}$.
Similarly,  Fig. \ref{fig:robustness} (b) shows the results of $\triangle ECE(\mu^i,\mu^j)$ for all methods.
One can see that our method (red) shows the best overall stability, regardless of the level of noise added.

Fig. \ref{fig:uncertainty_results} illustrates the segmentation results and uncertainty maps of different methods. 
Specifically, for \textbf{Uncertainty Interpretability}: (1) The second row overlays uncertainty maps with image gradients. Our method shows lower uncertainty in regions with higher gradients, conforming to \textbf{Principle 1}, while other methods lack this clear relationship. (2) The third row displays uncertainty difference maps between a test image and the image with patch noises. Our method effectively highlights noisy patches near edges, which conforms to \textbf{Principle 2}. 
For \textbf{Uncertainty Robustness}, the fourth row shows uncertainty difference maps between an original image and the noisy one with noise imposed on the entire image. 
Our method emphasizes edge regions while maintaining stable uncertainty in non-edge areas (\textbf{Principle 3}), while other methods might exhibit significant uncertainty variations in regions beyond the boundary.

\section{Conclusion}\label{sec:conclusion}
In this paper, we introduce a human-inspired uncertainty supervision method within the evidential learning framework. By utilizing image gradients and noise to constrain the uncertainty estimation, we not only provide reliable predictions but also offer interpretable and robust uncertainty estimations, which aligns with human experience. The proposed approach aids in understanding the sources of uncertainty, thereby facilitating better decision-making.

%
%

\begin{credits}
\subsubsection{\ackname} This work was funded by the National Natural Science Foundation of China (grant No. 62372115) and
Shanghai Municipal Education Commission-Artificial Intelligence Initiative to Promote Research Paradigm Reform and Empower Disciplinary Advancement Plan (grant no. 24KXZNA13).

\subsubsection{\discintname}
The authors have no competing interests to declare that are
relevant to the content of this article.
\end{credits}
%
%
%
\bibliographystyle{splncs04}
\bibliography{Paper-3770}

\begin{thebibliography}{10}
\providecommand{\url}[1]{\texttt{#1}}
\providecommand{\urlprefix}{URL }
\providecommand{\doi}[1]{https://doi.org/#1}

\bibitem{addimulam2020deep}
Addimulam, S., Mohammed, M.A., Karanam, R.K., Ying, D., Pydipalli, R., Patel, B., Shajahan, M.A., Dhameliya, N., Natakam, V.M.: Deep learning-enhanced image segmentation for medical diagnostics. Malaysian Journal of Medical and Biological Research  \textbf{7}(2),  145--152 (2020)

\bibitem{bernard2018deep}
Bernard, O., Lalande, A., Zotti, C., Cervenansky, F., Yang, X., Heng, P.A., Cetin, I., Lekadir, K., Camara, O., Ballester, M.A.G., et~al.: Deep learning techniques for automatic mri cardiac multi-structures segmentation and diagnosis: is the problem solved? IEEE transactions on medical imaging  \textbf{37}(11),  2514--2525 (2018)

\bibitem{cao2022swin}
Cao, H., Wang, Y., Chen, J., Jiang, D., Zhang, X., Tian, Q., Wang, M.: Swin-unet: Unet-like pure transformer for medical image segmentation. In: European conference on computer vision. pp. 205--218. Springer (2022)

\bibitem{erdur2024deep}
Erdur, A.C., Rusche, D., Scholz, D., Kiechle, J., Fischer, S., Llori{\'a}n-Salvador, {\'O}., Buchner, J.A., Nguyen, M.Q., Etzel, L., Weidner, J., et~al.: Deep learning for autosegmentation for radiotherapy treatment planning: State-of-the-art and novel perspectives. Strahlentherapie und Onkologie pp. 1--19 (2024)

\bibitem{eslami2018probabilistic}
Eslami, A., Paredes, B.R., Meyer, C., Rezende, D.J., De~Fauw, J., Ledsam, J., Maier-Hein, K.H., Ronneberger, O., Kohl, S.: A probabilistic u-net for segmentation of ambiguous images  (2018)

\bibitem{falk2019u}
Falk, T., Mai, D., Bensch, R., {\c{C}}i{\c{c}}ek, {\"O}., Abdulkadir, A., Marrakchi, Y., B{\"o}hm, A., Deubner, J., J{\"a}ckel, Z., Seiwald, K., et~al.: U-net: deep learning for cell counting, detection, and morphometry. Nature methods  \textbf{16}(1),  67--70 (2019)

\bibitem{gal2016dropout}
Gal, Y., Ghahramani, Z.: Dropout as a bayesian approximation: Representing model uncertainty in deep learning. In: international conference on machine learning. pp. 1050--1059. PMLR (2016)

\bibitem{guo2017calibration}
Guo, C., Pleiss, G., Sun, Y., Weinberger, K.Q.: On calibration of modern neural networks. In: International conference on machine learning. pp. 1321--1330. PMLR (2017)

\bibitem{huang2020unet}
Huang, H., Lin, L., Tong, R., Hu, H., Zhang, Q., Iwamoto, Y., Han, X., Chen, Y.W., Wu, J.: Unet 3+: A full-scale connected unet for medical image segmentation. In: ICASSP 2020-2020 IEEE international conference on acoustics, speech and signal processing (ICASSP). pp. 1055--1059. IEEE (2020)

\bibitem{huang2022lymphoma}
Huang, L., Ruan, S., Decazes, P., Den{\oe}ux, T.: Lymphoma segmentation from 3d pet-ct images using a deep evidential network. International Journal of Approximate Reasoning  \textbf{149},  39--60 (2022)

\bibitem{huttenlocher1993comparing}
Huttenlocher, D.P., Klanderman, G.A., Rucklidge, W.J.: Comparing images using the hausdorff distance. IEEE Transactions on pattern analysis and machine intelligence  \textbf{15}(9),  850--863 (1993)

\bibitem{josang2016subjective}
J{\o}sang, A.: Subjective logic, vol.~3. Springer (2016)

\bibitem{joshi2009multi}
Joshi, A.J., Porikli, F., Papanikolopoulos, N.: Multi-class active learning for image classification. In: 2009 ieee conference on computer vision and pattern recognition. pp. 2372--2379. IEEE (2009)

\bibitem{jungo2019assessing}
Jungo, A., Reyes, M.: Assessing reliability and challenges of uncertainty estimations for medical image segmentation. In: Medical Image Computing and Computer Assisted Intervention--MICCAI 2019: 22nd International Conference, Shenzhen, China, October 13--17, 2019, Proceedings, Part II 22. pp. 48--56. Springer (2019)

\bibitem{kendall2017uncertainties}
Kendall, A., Gal, Y.: What uncertainties do we need in bayesian deep learning for computer vision? Advances in neural information processing systems  \textbf{30} (2017)

\bibitem{krithika2022review}
Krithika Alias~AnbuDevi, M., Suganthi, K.: Review of semantic segmentation of medical images using modified architectures of unet. Diagnostics  \textbf{12}(12), ~3064 (2022)

\bibitem{lakshminarayanan2017simple}
Lakshminarayanan, B., Pritzel, A., Blundell, C.: Simple and scalable predictive uncertainty estimation using deep ensembles. Advances in neural information processing systems  \textbf{30} (2017)

\bibitem{mehrtash2020confidence}
Mehrtash, A., Wells, W.M., Tempany, C.M., Abolmaesumi, P., Kapur, T.: Confidence calibration and predictive uncertainty estimation for deep medical image segmentation. IEEE transactions on medical imaging  \textbf{39}(12),  3868--3878 (2020)

\bibitem{ORLANDO2020101570}
Orlando, J.I., Fu, H., Breda, J.B., van Keer, K., Bathula, D.R., Diaz-Pinto, A., Fang, R., Heng, P.A., Kim, J., Lee, J., et~al.: Refuge challenge: A unified framework for evaluating automated methods for glaucoma assessment from fundus photographs. Medical image analysis  \textbf{59},  101570 (2020)

\bibitem{ovadia2019can}
Ovadia, Y., Fertig, E., Ren, J., Nado, Z., Sculley, D., Nowozin, S., Dillon, J., Lakshminarayanan, B., Snoek, J.: Can you trust your model's uncertainty? evaluating predictive uncertainty under dataset shift. Advances in neural information processing systems  \textbf{32} (2019)

\bibitem{sensoy2018evidential}
Sensoy, M., Kaplan, L., Kandemir, M.: Evidential deep learning to quantify classification uncertainty. Advances in neural information processing systems  \textbf{31} (2018)

\bibitem{shafer1992dempster}
Shafer, G.: Dempster-shafer theory. Encyclopedia of artificial intelligence  \textbf{1},  330--331 (1992)

\bibitem{wang2019aleatoric}
Wang, G., Li, W., Aertsen, M., Deprest, J., Ourselin, S., Vercauteren, T.: Aleatoric uncertainty estimation with test-time augmentation for medical image segmentation with convolutional neural networks. Neurocomputing  \textbf{338},  34--45 (2019)

\bibitem{wissler1905spearman}
Wissler, C.: The spearman correlation formula. Science  \textbf{22}(558),  309--311 (1905)

\bibitem{yu2019uncertainty}
Yu, L., Wang, S., Li, X., Fu, C.W., Heng, P.A.: Uncertainty-aware self-ensembling model for semi-supervised 3d left atrium segmentation. In: Medical Image Computing and Computer Assisted Intervention--MICCAI 2019: 22nd International Conference, Shenzhen, China, October 13--17, 2019, Proceedings, Part II 22. pp. 605--613. Springer (2019)

\bibitem{zou2023towards}
Zou, K., Chen, Y., Huang, L., Yuan, X., Shen, X., Wang, M., Goh, R., Liu, Y., Fu, H.: Towards reliable medical image segmentation by utilizing evidential calibrated uncertainty. arXiv preprint arXiv:2301.00349  (2023)

\bibitem{zou2023evidencecap}
Zou, K., Yuan, X., Shen, X., Chen, Y., Wang, M., Goh, R.S.M., Liu, Y., Fu, H.: Evidencecap: towards trustworthy medical image segmentation via evidential identity cap. arXiv preprint arXiv:2301.00349  (2023)

\end{thebibliography}
%




\end{document}